\documentclass{article} %
\usepackage{iclr2027_conference,times}

\usepackage{amsmath,amsfonts,bm}

\def\eqref#1{equation~\ref{#1}}

\def\1{\bm{1}}

\DeclareMathAlphabet{\mathsfit}{\encodingdefault}{\sfdefault}{m}{sl}
\SetMathAlphabet{\mathsfit}{bold}{\encodingdefault}{\sfdefault}{bx}{n}

\usepackage{hyperref}
\usepackage{url}

\usepackage{microtype}
\usepackage{graphicx}
\usepackage{wrapfig}
\usepackage{subcaption}
\usepackage{booktabs} %

\usepackage[capitalize,noabbrev]{cleveref}
\usepackage{colortbl}
\usepackage[table]{xcolor}
\usepackage{multirow}
\usepackage{multicol}
\graphicspath{{Figure/}}
\usepackage{enumitem}
\usepackage[american]{babel}
\usepackage{nicefrac}       %
\usepackage{bbding}
\usepackage{arydshln}

\definecolor{secondcolor}{RGB}{180,198,231}
\definecolor{firstcolor}{RGB}{240,220,220}

\newcommand{\bfsection}[1]{\vspace*{0.00cm}\noindent\textbf{#1}.}

\usepackage{xspace}
\newcommand{\eg}{\emph{e.g.}\xspace}

\newcommand{\ie}{\emph{i.e.}\xspace}

\newcommand{\bgfirst}{\cellcolor[rgb]{.941, .863, .863}}
\newcommand{\bgsecond}{\cellcolor[rgb]{.706,  .776,  .906}}

\crefname{section}{Sec.}{Secs.}
\Crefname{section}{Section}{Sections}
\crefname{figure}{Fig.}{Figs.}
\Crefname{figure}{Figure}{Figures}
\Crefname{table}{Table}{Tables}
\crefname{table}{Tab.}{Tabs.}
\crefname{equation}{Eq.}{Eqs.}
\Crefname{equation}{Equation}{Equations}

\title{MambaVF: State Space Model for Efficient Video Fusion}

\author{Zixiang Zhao$^{1}$\quad
Yukun Cui$^{2}$\thanks{Corresponding author.}\quad
Lilun Deng$^{2}$\quad
Haowen Bai$^{3}$\\
\textbf{Haotong Qin$^{4}$\quad
Tao Feng$^{5}$\quad
Konrad Schindler$^{1}$}\\[1mm]
$^{1}$ETH Z\"urich\quad
$^{2}$Xi’an Jiaotong University\quad
$^{3}$Nanyang Technological University\\
$^{4}$Hong Kong Polytechnic University\quad
$^{5}$Tsinghua University\\[1mm]
\texttt{zixiang.zhao@ethz.ch}
}

\iclrfinalcopy %
\begin{document}

\maketitle

\begin{abstract}
Video fusion aims to integrate complementary information from multiple source videos while preserving temporal consistency. 
Effective modeling of temporal dynamics is essential to this goal, yet existing methods incur substantial computational overhead from optical flow estimation and feature warping.
In this paper, we present \textbf{MambaVF}, an efficient video fusion framework that uses state space model (SSM) to achieve temporal modeling without explicit motion estimation. First, by formulating video fusion as a sequential state update process, MambaVF captures long-range temporal dependencies with linear complexity, significantly reducing computation and memory costs. Second, the lightweight SSM-based fusion module eliminates conventional flow-guided alignment. Instead, it introduces a mutual state fusion module and a spatio-temporal bidirectional scanning mechanism to enable information aggregation across video streams. Experiments on multiple benchmarks confirm that MambaVF reaches state-of-the-art performance in different video fusion applications (multi-exposure, multi-focus, infrared-visible, medical), while reducing parameters by \textgreater90\% and FLOPs by \textgreater80\%, resulting in \textgreater50\% shorter runtime. 
Project page: \href{https://mambavf.github.io}{\textit{mambavf.github.io}}.
\end{abstract}
\begin{wrapfigure}{r}{0.53\textwidth}
  \centering
  \vspace{-1em}
  \includegraphics[width=\linewidth]{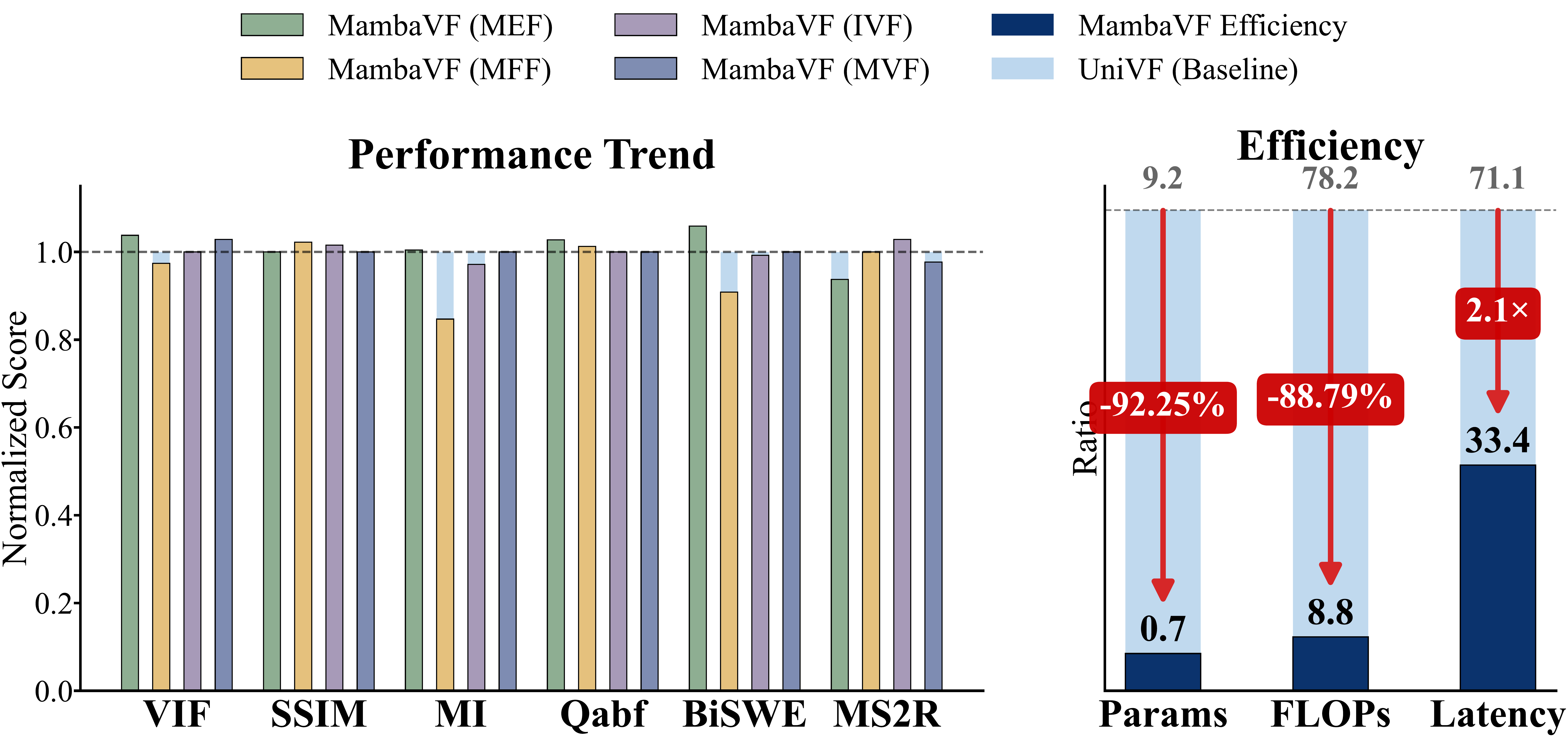}
  \vspace{-0.5em}
    \caption{Compared with UniVF \citep{zhao2025unified}, our MambaVF not only attains state-of-the-art performance on VF-Bench \citep{zhao2025unified}, but requires only \textbf{5.7\%} of the parameters and \textbf{13.4\%} of the FLOPs, while achieving a \textbf{2.1×} speedup.}
  \label{fig:first}
  \vspace{-1em}
\end{wrapfigure}

\section{Introduction}\label{sec:intro}
Video fusion aims to integrate complementary information from different sensors (\eg, infrared and visible, or CT and MRI) or acquisition conditions (\eg, different exposure times or focus settings) into a unified video sequence~\citep{zhao2025unified}.
The enriched output video can significantly enhance video interpretation~\citep{Zhao_2023_CVPR,DBLP:conf/cvpr/LiuFHWLZL22,zhang2021deep}. By mitigating sensing limitations and degradations in the individual inputs, fusion produces more informative videos for applications including  computational imaging~\citep{Chen_2025_CVPR}, autonomous driving~\citep{prakash2021multi}, robotic perception~\citep{li2023see}, and clinical diagnosis~\citep{10190200}.

\begin{wrapfigure}{r}{0.48\textwidth}
  \centering
  \vspace{-1em}
  \includegraphics[width=\linewidth]{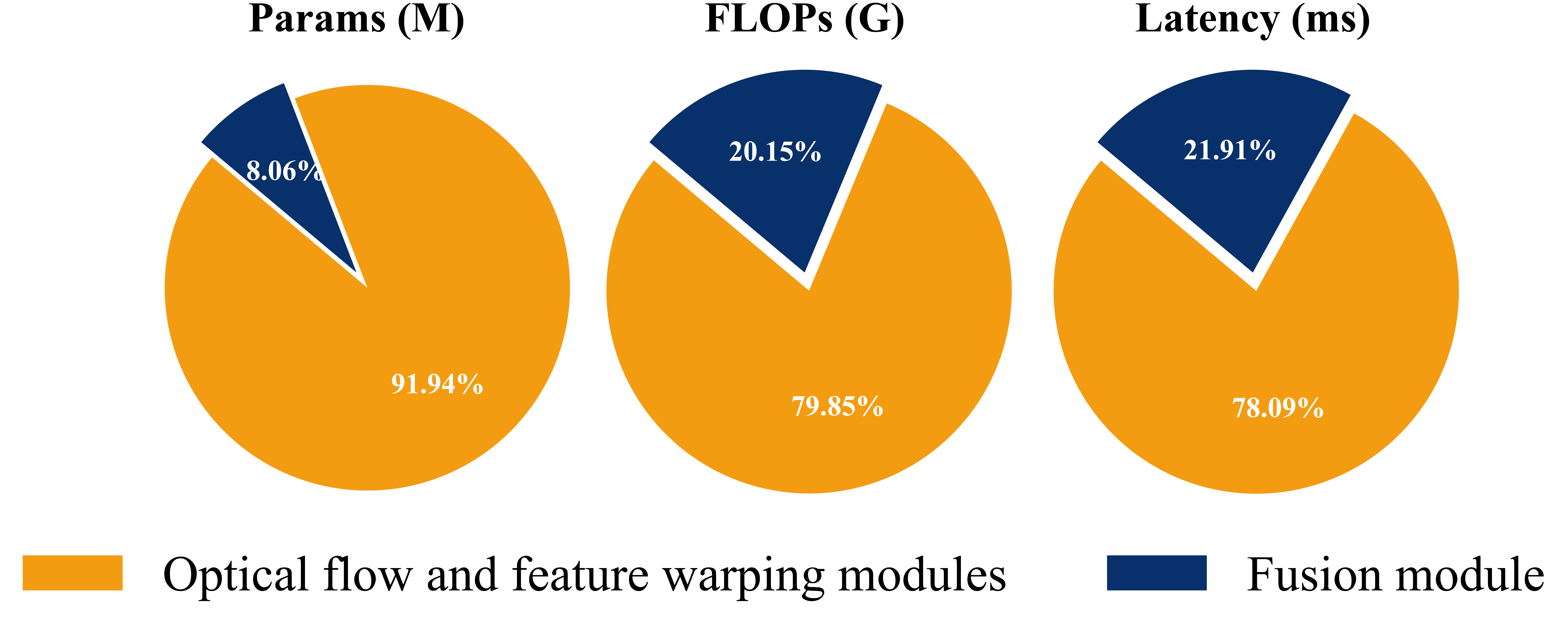}
  \vspace{-0.5em}
  \caption{Parameter, FLOPs, and runtime contribution of optical flow and feature warping modules in UniVF~\citep{zhao2025unified}.}
  \label{fig:flow_cost_in_univf}
  \vspace{-.5em}
\end{wrapfigure}

Existing video fusion methods typically require explicit optical flow estimation between consecutive frames, followed by feature warping, to aggregate information~\citep{zhao2025unified}.
Such a flow-based approach faces two fundamental bottlenecks. 
(i) \textit{Computational cost:} optical flow estimation and warping are expensive. As illustrated in \cref{fig:flow_cost_in_univf} for UniVF~\citep{zhao2025unified}, they account for $\approx$80\% of the total runtime and FLOPs, and $\approx$92\% of all parameters. 
(ii) \textit{Sensitivity to flow errors:} 
the quality of fusion is upper-bounded by the correctness of the estimated flow field. Occlusions, fast motion, or illumination changes that degrade the flow directly cause blur and ghosting artifacts in the fusion result.
Conceptually, Transformers would make it possible to globally and robustly aggregate spatio-temporal information across frames via self-attention. But that would incur an even higher computational burden, due to the quadratic complexity of the attention mechanism.
The high computational cost renders naive Transformer architectures impractical for video fusion, especially since applications like infrared-visible, multi-exposure and multi-focus fusion should often be deployed on-device, \ie, on edge hardware and in near real-time.

Meanwhile, it has been shown that state space models (SSMs) such as the Mamba architecture~\citep{gu2024mamba} are often able to capture long-range dependencies almost as well as classical Transformers, but with linear computational complexity. Mamba’s efficient latent-space updating mechanism is a natural fit for video sequences, a data type characterized by high temporal redundancy and long-term correlations~\citep{li2024videomamba,argaw2025mambavideo}.
Of particular interest for our problem is that Mamba enables feature extraction in space and time without explicit, computationally heavy alignment and warping modules based on optical flow.

Motivated by this insight, we propose \textbf{MambaVF}, an efficient, SSM-driven video fusion framework. It departs from the usual practice of explicitly computing optical flow for frame-to-frame alignment.
Instead, it leverages a \textit{mutual state fusion} (MuFuse) module that captures the temporal dynamics of both input videos in a unified manner, enabling interaction between their hidden states.
To ensure the model can comprehensively represent complex video content, we complement  MuFuse with a \textit{spatio-temporal bidirectional} (STB) scanning mechanism.
Compared to standard 1D or 2D scanning, it performs a comprehensive traversal across spatial coordinates and frame indices.
STB scanning improves spatio-temporal awareness in the latent representation, to ensure the model preserves the most informative structures from both inputs and outputs a high-fidelity fusion result.

In summary, the contributions of our work are:
\begin{itemize}[itemsep=2pt, topsep=0pt, parsep=0pt]
    \item \textbf{MambaVF}, to our knowledge the first video fusion framework based on SSMs. By eliminating the need for optical flow computation and warping, it offers a new, highly efficient approach to video fusion.
    \item The \textbf{MuFuse} Mamba Encoder and associated \textbf{STB} scanning mechanism, which extend the standard Mamba block towards holistic video understanding across two (or more) different, complementary source streams.
    \item An extensive evaluation on VF-Bench, in which MambaVF matches the state-of-the-art performance of UniVF, while using only \textbf{5.7\%} of the parameters and \textbf{13.4\%} of the FLOPs, and running \textbf{2.1×} faster.
\end{itemize}

\section{Related Work}\label{sec:2}
\bfsection{Image and Video Fusion}
Deep learning-based image fusion~\citep{Zhao_2023_CVPR,zhang2021deep} can be broadly divided into discriminative and generative paradigms. Discriminative methods~\citep{DBLP:conf/cvpr/Xu0ZSL021,DBLP:journals/tcsv/ZhaoXZLZL22,li2023lrrnet,DBLP:conf/cvpr/ZhaoZXLP22} use CNNs~\citep{prabhakar2017deepfuse,han2022multi,DBLP:journals/tip/GaoDXXD22,bouzos2023convolutional,guan2023mutual,deng2021deep} and Transformers~\citep{qu2022transmef,guan2023mutual} via model- or data-driven strategies~\citep{DBLP:conf/cvpr/Xu0ZSL021,DBLP:journals/tcsv/ZhaoXZLZL22,li2023lrrnet,DBLP:conf/cvpr/ZhaoZXLP22,zhaoijcai2020,Zhao_2023_CVPR,li2018densefuse}. Generative approaches~\citep{ma2019fusiongan,DBLP:conf/cvpr/LiuFHWLZL22} treat fusion as distribution fitting, evolving from GANs~\citep{ma2019fusiongan,DBLP:journals/tip/MaXJMZ20} to Diffusion models~\citep{Chen_2025_CVPR,Zhao_2023_ICCV}, and recently to efficient one-step generation via rectified flow \citep{wang2025efficient}.
To obtain more robust results, methods employ registration~\citep{DBLP:C2RF,DBLP:conf/ijcai/WangLFL22,DBLP:conf/mm/JiangZ0L22,DBLP:conf/cvpr/Xu0YLL22} for misalignment and combined restoration for complex degradations~\citep{Cao_2025_ICCV,10994384}. Concurrently, fusion is guided by downstream task feedback~\citep{Liu_2025_CVPR_DcEVO,DBLP:journals/tcsv/BaiZZJDCXZ25,Zhong_2025_ICCV}, cross-task synergy~\citep{DBLP:GIFNet,9151265,Liang2022ECCV,DBLP:TCMOA}, and meta-learning~\citep{DBLP:ReFusion,Bai_2025_CVPR,DBLP:conf/cvpr/ZhaoXZHL23}. Recently, vision-language priors ~\citep{DBLP:FILM} and text prompts ~\citep{DBLP:Text-Diffuse}  have emerged for interactive, controllable fusion.

Then, to enhance the understanding of dynamics, image fusion has been extended to the video domain. 
\citet{xie2024rcvs,tang2026videofusion,gong2025temcoco,Liu_2026_CVPR,li2026drfusion,li2026mavfusion} mainly focus on multi-modal scenarios. \citet{zhao2025unified} propose UniVF, a unified video fusion framework that provides a comprehensive multi-frame learning paradigm for multiple fusion tasks, including multi-exposure, multi-focus, infrared-visible, and medical video fusion. This framework standardizes the learning objective, datasets, and evaluation protocols across these diverse scenarios.

\bfsection{State Space Models in Vision}
State space models~\citep{gu2021efficiently} provide an efficient basis for sequence modeling. In particular, Mamba~\citep{gu2024mamba} has shown strong performance in natural language processing by introducing a selective scan mechanism. Mamba demonstrates powerful capabilities in modeling long-range dependencies with linear computational complexity, bridging the gap between the efficiency of CNNs and the global receptive field of Transformers. Given the success of these efficient designs, many studies have adapted 1D SSMs to 2D visual tasks. Architectures like Vim~\citep{zhu2024vision} and VMamba~\citep{liu2024vmamba} use bidirectional or cross-scanning strategies to process images in visual recognition. In low-level vision, MambaIR~\citep{guo2024mambair} utilizes SSMs to restore pixel-level image details. Furthermore, recent works like VideoMamba~\citep{li2024videomamba}, Vivim~\citep{yang2024vivim}, Vamba~\citep{Ren_2025_ICCV} and Video Mamba Suite~\citep{chen2026video} have extended this to the spatiotemporal domain, providing an efficient way to maintain temporal consistency without substantial computational overhead. Inspired by this, we aim to introduce the Mamba architecture to multi-source video fusion tasks, utilizing its efficient global modeling ability to handle cross-modal interactions and temporal evolution simultaneously.

\section{MambaVF}
\label{sec:method}
In this section, we present MambaVF, an efficient state space model for unified video fusion. We first outline the framework, then describe the \textit{Mutual State Fusion} (MuFuse) Mamba Encoder, \textit{Spatio-Temporal Bidirectional} (STB) Scanning mechanism, and frame reconstruction.
\subsection{Overview}

Given synchronized source video windows $S_1, S_2 \in \mathbb{R}^{B \times T \times C \times H \times W}$, where $B, T, C, H, W$ denote the batch size, number of frames, channels, height, and width, respectively, MambaVF generates a fused video sequence that integrates complementary information without explicit motion estimation. 

\begin{figure*}[t]
    \centering
    \includegraphics[width=\linewidth]{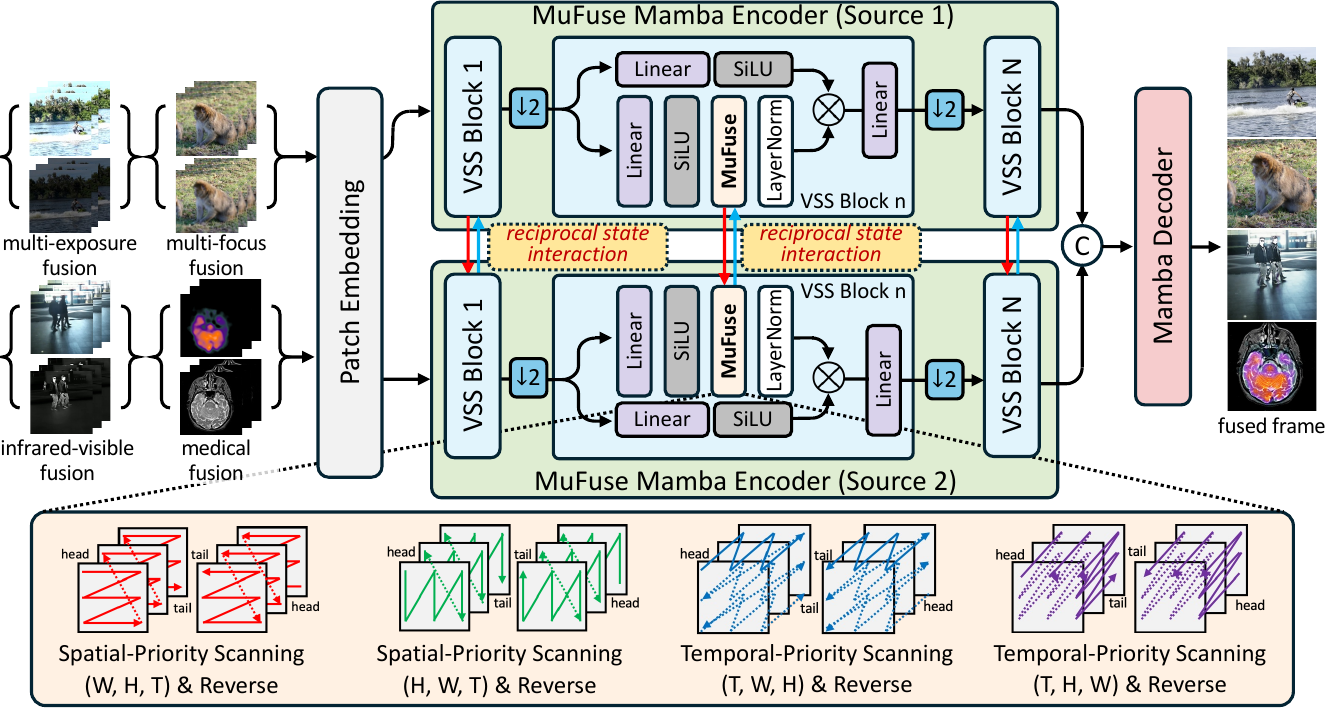}
    \caption{An overview of our \textbf{MambaVF} architecture. The MuFuse Mamba Encoder processes the embedded source features through VSS Blocks, whose MuFuse modules exchange complementary information via reciprocal state interaction. STB scanning supplies eight matching paths for the two sources: four forward orders with their full reversals. Tuples list axes from fastest to slowest. Spatial downsampling reduces encoding cost, and the decoder reconstructs the fused center frame.}
    \label{fig:pipeline}
\end{figure*}

The overall pipeline consists of three main stages, as illustrated in \cref{fig:pipeline}. First, the input videos are processed by a shared patch embedding layer to map raw pixels into a latent embedding space. 
Second, a dual-stream MuFuse Mamba Encoder jointly models temporal dependencies and cross-source interactions, with spatial downsampling to reduce computation. Third, the two streams are concatenated and passed to a Mamba Decoder with PixelShuffle upsampling to reconstruct the central frame of each input window.
This multiscale Mamba encoder-decoder aggregates spatio-temporal context through hidden states and reconstructs the fused frame without flow-guided alignment.

\subsection{Feature Extraction and Interaction}

\bfsection{Feature Embedding}
MambaVF utilizes a dual-stream architecture to preserve the distinct characteristics of each source (\eg, the thermal radiation in infrared and the texture details in visible videos). The two streams share a patch embedding layer:
\begin{equation}
    E_{S}^1 = \text{Embedding}(S_1), \quad
    E_{S}^2 = \text{Embedding}(S_2),
\end{equation}
producing a spatio-temporal feature volume for each source.
The MuFuse Mamba Encoder processes these features through \textit{Visual State Space} (VSS) Blocks, each containing a MuFuse Module that operates on sequences constructed by STB Scanning. Selective scanning scales linearly with the token count for fixed feature and state dimensions.

\bfsection{MuFuse Mamba Encoder}
For a single source or modality, Mamba~\citep{gu2024mamba} scans a token sequence by updating a hidden state and computing contextual outputs as follows:
\begin{equation}
h_t=\bar{A}_t h_{t-1}+\bar{B}_t x_t, \quad
y_t=C_t h_t+D x_t.
\end{equation}
Here, $\bar{A}_t$ and $\bar{B}_t$ govern state propagation and input injection, while $C_t$ maps the state to the output and $D$ provides a direct input contribution. Mamba conditions $B_t$, $C_t$, and the discretization step size $\Delta_t$ on the current input, resulting in input-adaptive state updates.

Video fusion hinges on two equally important capabilities: extracting distinctive spatio-temporal cues from each source and exchanging complementary context between sources. The MuFuse Mamba Encoder therefore couples source-specific state modeling with reciprocal state interaction, so that each stream preserves its own contextual representation while benefiting from the other source's accumulated information.
Within each VSS Block in the encoder, the two streams are projected into scan features and gating branches. The scan features follow matching STB paths and are processed by the MuFuse Module. The resulting outputs are restored, normalized, modulated by the corresponding source’s SiLU gate, and projected back to the original feature width.

The MuFuse Module implements this interaction by combining source-specific state evolution with reciprocal state fusion. Along each pair of matching one-dimensional STB sequences (detailed below), each source independently accumulates context in its hidden state:
\begin{equation}
    h_t^m = \bar{A}_t^{m}h_{t-1}^m + \bar{B}_t^{m}x_t^m,\quad m\in\{1,2\},
    \label{eq:mufuse_state}
\end{equation}
where $t$ indexes scan positions and $x_t^m$ is a projected source token. The states are coupled at the output through an input-dependent block projection:
\begin{equation}
\begin{bmatrix}y_t^1\\y_t^2\end{bmatrix}
=
\underbrace{\begin{bmatrix}
C_{1,t} & C_{2,t}\\
C_{3,t} & C_{4,t}
\end{bmatrix}}_{\mathcal{C}_t:\ \text{reciprocal state fusion}}
\begin{bmatrix}h_t^1\\h_t^2\end{bmatrix}
+
\begin{bmatrix}D_1x_t^1\\D_2x_t^2\end{bmatrix}.
\label{eq:mufuse_readout}
\end{equation}
The diagonal blocks preserve source-specific state readout, while the off-diagonal blocks enable reciprocal cross-source readout, allowing each output to incorporate the other source's state context. {$C_{1,t}$ and $C_{3,t}$ are generated from $x_t^1$, whereas $C_{2,t}$ and $C_{4,t}$ are generated from $x_t^2$, using four distinct learned projections.}
Each state summarizes a scan history, so both streams receive contextual information from both sources while retaining separate within-block state dynamics. The interaction-enhanced outputs feed subsequent VSS Blocks.

The two entries in each column of $\mathcal{C}_t$ act on the same source state. A direct implementation nevertheless makes four scan calls, recomputing each source recurrence for self and cross outputs. We extend Mamba's low-level selective-scan operator through its C++ interface and CUDA forward/backward kernels to reuse each recurrence across both outputs, reducing four calls to two. This mathematically equivalent optimization reduces redundant computation while preserving numerical accuracy, as verified by FP32 forward and backward tests.

Finally, each VSS block is followed by $2\times$ spatial average pooling, which preserves temporal resolution while reducing the sequence length at the next stage to one quarter. This lowers the computational cost of context extraction and cross-source interaction.

\bfsection{Spatio-Temporal Bidirectional (STB) Scanning}
STB Scanning organizes video features into complementary spatial-priority and temporal-priority sequences for the MuFuse Module. Each of four forward orders is paired with its full reversal, yielding eight directions as illustrated in \cref{fig:pipeline}.

Let $x^m\in\mathbb{R}^{d\times T\times H_\ell\times W_\ell}$ denote a source feature volume at encoder stage $\ell$, where $d$ is the feature width. Axis tuples are ordered from fastest to slowest. The paths are organized as follows:
\begin{enumerate}[itemsep=2pt, topsep=0pt, parsep=0pt]
    \item \textit{Spatial-Priority Scanning}: $(W,H,T)$ traverses each frame row by row, whereas $(H,W,T)$ traverses it column by column, before advancing in time. These orders make horizontal or vertical neighbors adjacent in the sequence, supporting intra-frame texture and structure aggregation.
    \item \textit{Temporal-Priority Scanning}: $(T,W,H)$ and $(T,H,W)$ visit all frames at a fixed spatial location before advancing to the next location in row-major or column-major order. They reduce the sequence distance between adjacent frames at the same coordinate from $H_\ell W_\ell$ in the spatial-priority orders to one step, capturing temporal variation without explicitly tracking motion trajectories.
    \item \textit{Symmetrical Bidirectional Traversal}: Each forward path is paired with the reverse of its entire sequence. The two sources follow matching paths, so their states summarize corresponding scan histories for reciprocal state interaction.
\end{enumerate}
After the multi-path scanning, the eight sequences are mapped back to their original spatio-temporal dimensions and integrated to form the enhanced feature representation.
Existing Mamba variants explore adaptive spatial ordering and multiple spatiotemporal scanning routes \citep{li2026damamba,feng2025highresolution}, while interaction-oriented designs organize scans across tasks or interleave modalities along spatial directions \citep{cao2025bim,wang2026mmmamba}.
Our MuFuse Module combines eight spatiotemporal scanning paths with complementary spatial and temporal priorities, applying the same traversal order to both sources along each path.
The distinction lies in coupling shared spatiotemporal traversal with state interaction. Each scanning path models context within each source while exchanging information between the two sources.

\subsection{Fusion and Reconstruction}
The MuFuse Mamba Encoder $\mathcal{E}_{\mathrm{MuFuse}}$ produces two feature volumes $F_1,F_2$, which are merged by channel-wise concatenation:
\begin{equation}
F_1, F_2  = \mathcal{E}_{\mathrm{MuFuse}}(E_{S}^1,E_{S}^2), \ 
F_{\text{fused}} = \text{Concat}([F_1, F_2]).
\end{equation}
The fused features $F_{\text{fused}}$ are subsequently processed by the Mamba Decoder, which applies three stages of $2\times$ PixelShuffle upsampling with spatial Mamba refinement. 
Accordingly, it uses four-way spatial scanning for center-frame reconstruction. 
This design choice concentrates the state-space modeling on intra-frame structural refinement, ensuring high-fidelity spatial reconstruction for the target frame. 
Finally, 2D residual blocks~\citep{HeZRS16} are lastly used to restore fine-grained textures and project the high-dimensional embeddings back to the pixel domain. 

\section{Experiments}\label{sec:exp}
MambaVF is evaluated on VF-Bench~\citep{zhao2025unified} across four diverse scenarios: \textit{multi-exposure fusion} (MEF), \textit{multi-focus fusion} (MFF), \textit{infrared-visible fusion} (IVF), and \textit{medical video fusion} (MVF). We report experimental setup, fusion quality, computational efficiency, and ablation studies.
\begin{figure*}[t]
  \vspace{-2em}
  \centering
  \includegraphics[width=1.0\linewidth]{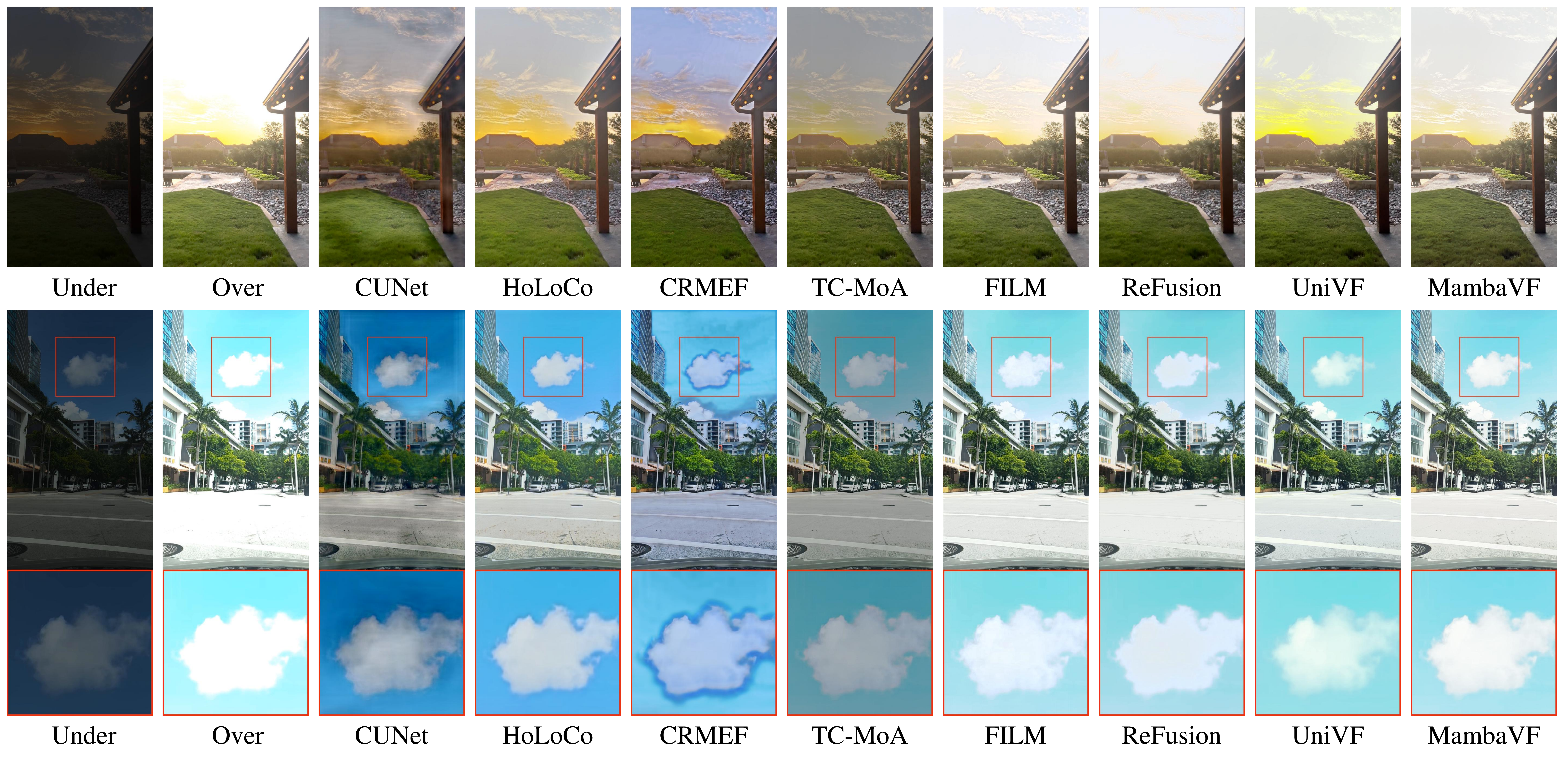}
  \caption{Visual comparison of fused results for multi-exposure video fusion.}
  \label{fig:MEF}
  \vspace{-1em}
\end{figure*}

\begin{figure*}[t]
  \centering
  \includegraphics[width=1.0\linewidth]{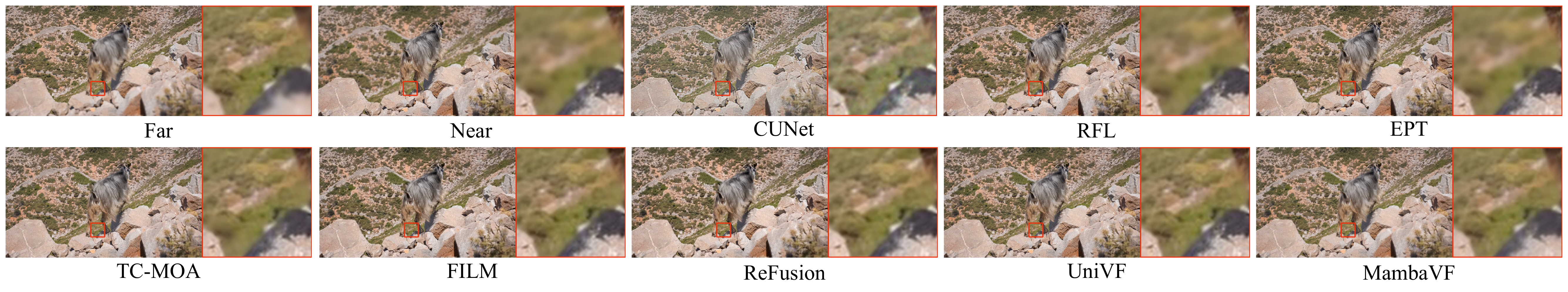}
  \caption{Visual comparison of fused results for multi-focus video fusion.}
  \label{fig:MFF}
  \vspace{-1em}
\end{figure*}

\subsection{Experimental Configuration}

\bfsection{Loss Function}
Following the training setup of UniVF~\citep{zhao2025unified}, MambaVF is trained using a multi-term loss function that balances spatial, structural, and temporal fidelity:
\begin{equation}\label{eq:loss_total}
    \mathcal{L} = \mathcal{L}_{\text{spatial}} + \alpha_1 \mathcal{L}_{\text{grad}} + \alpha_2 \mathcal{L}_{\text{temp}}.
\end{equation}
The individual loss components are as follows:

(a) \textit{Spatial Similarity} ($\mathcal{L}_{\text{spatial}}$): This term supervises the photometric distribution of the fused output. For IVF and MVF, we maximize intensity saliency: $\mathcal{L}_{\text{spatial}} = \frac{1}{HW} \|I^F_t - \max(I^1_t, I^2_t)\|_1$. For MEF, it combines a mean intensity term $\mathcal{L}_{int}\!=\!\frac{1}{HW}\!\|I^F_t\!- \text{mean}(I^1_t, I^2_t)\|_1$ and the MEF-SSIM loss \citep{ma2015perceptual}. For MFF, $\mathcal{L}_{\text{spatial}}$ is equivalent to $\mathcal{L}_{int} = \frac{1}{HW} \|I^F_t - \text{mean}(I^1_t, I^2_t)\|_1$.

(b) \textit{Gradient Preservation} ($\mathcal{L}_{\text{grad}}$): To sharpen fused textures, we utilize a Sobel-based gradient loss: $\mathcal{L}_{\text{grad}} = \frac{1}{HW} \| |\nabla I^F_t| - \max(|\nabla I^1_t|, |\nabla I^2_t|) \|_1$, encouraging the retention of high-frequency structural details from both sources.

(c) \textit{Temporal Consistency} ($\mathcal{L}_{\text{temp}}$): This term penalizes abrupt inter-frame intensity fluctuations, ensuring smooth photometric transitions without explicit motion compensation. For a detailed derivation of these objectives and settings, please refer to \citet{zhao2025unified}.

\bfsection{Implementation Details}
The MambaVF framework is implemented using PyTorch and trained on 4$\times$NVIDIA GH200 GPUs. We utilize the Adam optimizer with an initial learning rate of $1\times10^{-4}$, which is gradually reduced through an exponential decay strategy over 20k iterations. The training process uses a batch size of 32 with gradient accumulation. 
The MuFuse Mamba Encoder has three VSS Blocks with 32 feature channels per source.

\bfsection{Evaluation Protocol}
Following the evaluation protocol in VF-Bench~\citep{zhao2025unified}, our evaluation encompasses both spatial and temporal dimensions. For spatial quality, we employ \textit{Visual Information Fidelity} (VIF), \textit{Structural Similarity} (SSIM), \textit{Mutual Information} (MI), and $Q^{AB/F}$, assessing perceptual fidelity, structural similarity, shared information, and edge preservation, respectively. For temporal stability, we adopt \textit{Bi-Directional Self-Warping Error} (BiSWE) and \textit{Motion Smoothness with Dual Reference Videos} (MS2R) to measure inter-frame warping consistency and agreement with source-video flow changes, respectively. Higher values in spatial metrics and lower values in temporal metrics indicate superior fusion performance.

\begin{figure*}[t]
  \centering
  \includegraphics[width=\linewidth]{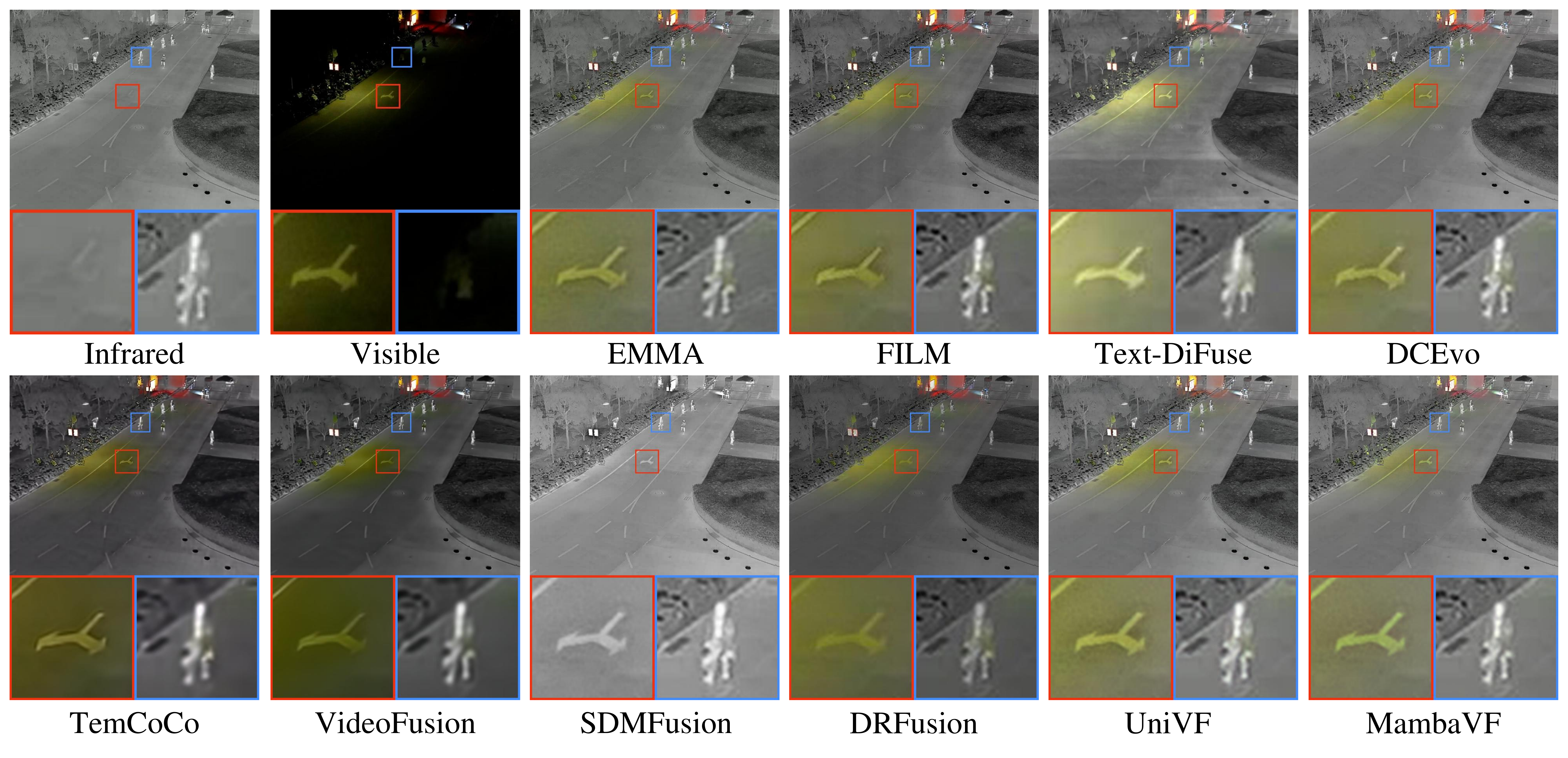}
  \caption{Visual comparison of fused results for infrared-visible video fusion.}
  \label{fig:IVF}
  \vspace{-1em}
\end{figure*}
\begin{figure}[t]
  \centering
  \includegraphics[width=\linewidth]{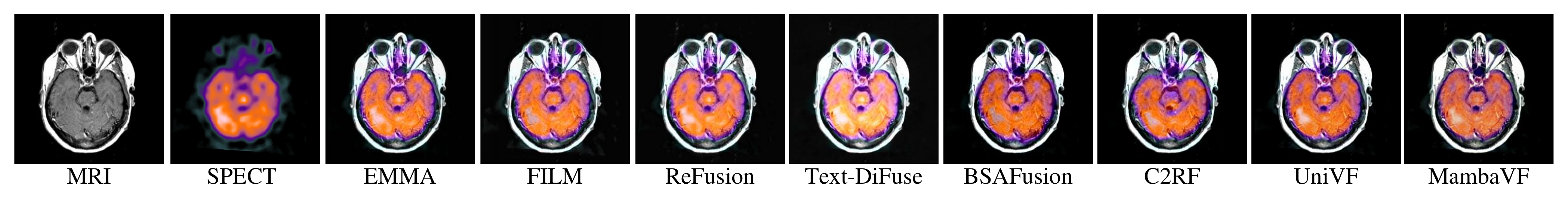}
  \caption{Visual comparison of fused results for MVF.}
  \label{fig:MVF}
  \vspace{-1em}
\end{figure}
\begin{table*}[t]
  \centering
  \caption{Quantitative evaluation results for the low-resolution MEF and MFF tasks. The \colorbox{firstcolor}{red} and \colorbox{secondcolor}{blue} highlights indicate the best and second-best scores.}
    \resizebox{\linewidth}{!}{
    \begin{tabular}{@{}cccccccccccccc@{}}
    \toprule
          & \multicolumn{6}{c}{\textbf{VF-Bench Multi-Exposure Fusion Branch (540p)}} & & \multicolumn{6}{c}{\textbf{VF-Bench Multi-Focus Fusion Branch (480p)}} \\
    \cmidrule(r){2-7} \cmidrule(l){9-14} %
     & VIF$\uparrow$ & SSIM$\uparrow$ & MI$\uparrow$ & Qabf$\uparrow$ & BiSWE$\downarrow$ & MS2R$\downarrow$ &  & VIF$\uparrow$ & SSIM$\uparrow$ & MI$\uparrow$ & Qabf$\uparrow$ & BiSWE$\downarrow$ & MS2R$\downarrow$ \\ 
    \midrule
    CUNet  & 0.50 & 0.85 & 1.85 & 0.39 & 7.55 & 0.20 & CUNet  & 0.53 & 0.86 & 3.52 & 0.68 & 10.23 & 0.42 \\
    HoLoCo & 0.50 & 0.86 & 2.56 & 0.42 & 8.22 & 0.19 & RFL    & \bgsecond{0.77} & 0.90 & 6.31 & 0.78 & 8.46  & 0.28 \\
    CRMEF  & 0.62 & 0.94 & 2.60 & 0.63 & 8.72 & 0.19 & EPT    & 0.76 & 0.90 & \bgsecond{6.33} & 0.78 & 8.50  & 0.29 \\
    TC-MoA & 0.74 & 0.99 & 2.93 & 0.72 & 7.82 & 0.16 & TC-MoA & 0.75 & 0.90 & 5.27 & 0.77 & 8.39  & 0.28 \\
    FILM   & 0.77 & 0.99 & 4.35 & 0.72 & 8.28 & 0.17 & FILM   & 0.75 & 0.89 & 5.06 & 0.78 & 8.61  & 0.33 \\
    ReFus  & 0.74 & 0.97 & 3.81 & 0.72 & 7.63 & 0.16 & ReFus  & 0.73 & 0.90 & 4.93 & 0.77 & \bgsecond{8.00}  & 0.32 \\
    UniVF   & \bgsecond{0.79} & \bgsecond{0.99} & \bgsecond{4.38} & \bgsecond{0.73} & \bgfirst{6.96} & \bgsecond{0.16} & UniVF   & \bgfirst{0.77} & \bgsecond{0.90} & \bgfirst{6.34} & \bgsecond{0.79} & 8.29  & \bgsecond{0.27} \\
    MambaVF & \bgfirst{0.82} &  \bgfirst{0.99} & \bgfirst{4.45} & \bgfirst{0.75} & \bgsecond{7.47} & \bgfirst{0.15} & MambaVF & 0.75 & \bgfirst{0.92} & 5.37 & \bgfirst{0.80} & \bgfirst{7.57} & \bgfirst{0.26} \\
    \bottomrule
    \end{tabular}}%
  \label{tab:MFFMEF}%
  \vspace{-0.5em}
\end{table*}%
\begin{table*}[t]
    \begin{minipage}[t]{0.47\textwidth}
        \centering
        \caption{Quantitative evaluation for IVF.}
        \label{tab:IVF}
        \resizebox{\linewidth}{!}{%
            \begin{tabular}{ccccccc}
            \toprule
            & \multicolumn{6}{c}{\textbf{VF-Bench Infrared-Visible Video Fusion Branch}} \\
            \cmidrule(lr){2-7} 
                  & VIF $\uparrow$   & SSIM $\uparrow$   & MI $\uparrow$    & Qabf $\uparrow$   & BiSWE $\downarrow$  & MS2R $\downarrow$  \\
            \midrule
            EMMA  & 0.37  & 0.63  & 2.01  & 0.58  & 4.79  & 0.37 \\
            FILM  & 0.40  & 0.63  & 2.05  & 0.64  & 4.78  & 0.37 \\
            Text-D & 0.30  & 0.60  & 1.64  & 0.39  & 10.63 & 0.40 \\
            DCEvo & 0.43  & 0.64  & {2.44} & 0.66  & 4.57  & 0.37 \\
            TemCoCo & 0.29 & 0.60 & 1.77 & 0.41 & 4.66 & 0.46 \\
            VideoFusion & 0.27 & 0.60 & 1.83 & 0.37 & \bgfirst{3.00} & 0.57 \\
            SDMFusion & 0.43 & 0.61 & 1.94 & 0.65 & 4.49 & 0.37 \\
            DRFusion & 0.35 & 0.64 & 1.98 & 0.53 & 4.90 & 0.38 \\
            UniVF  & \bgsecond{0.44} & \bgsecond{0.64} & \bgsecond{2.47} & \bgsecond{0.68} & {3.94} & \bgsecond{0.35} \\
            MambaVF & \bgfirst{0.45} & \bgfirst{0.65} & \bgfirst{2.51} & \bgfirst{0.68} & \bgsecond{3.89} & \bgfirst{0.35}\\
            \bottomrule
            \end{tabular}%
        }
    \end{minipage}
\hfill 
    \begin{minipage}[t]{0.535\textwidth}
        \centering
        \caption{Quantitative evaluation for MVF.}
        \label{tab:MVF}
        \resizebox{\linewidth}{!}{%
            \begin{tabular}{ccccccc}
            \toprule
            & \multicolumn{6}{c}{\textbf{VF-Bench Medical Video Fusion Branch}} \\
            \cmidrule(lr){2-7} 
                  & VIF $\uparrow$   & SSIM $\uparrow$   & MI $\uparrow$    & Qabf $\uparrow$   & BiSWE $\downarrow$  & MS2R $\downarrow$  \\
            \midrule
            EMMA  & 0.29  & 0.68  & 1.73  & 0.60  & 30.00 & 1.98 \\
            FILM  & 0.33 & 0.36  & 1.83 & 0.67  & 32.04 & 1.59 \\
            ReFus & 0.31  & 0.32  & 1.74  & 0.67 & 32.85 & 1.74 \\
            Text-D & 0.24  & 0.21  & 1.58  & 0.52  & 34.09 & 1.96 \\
            BSAF & 0.28  & 0.63  & 1.69  & 0.58  & 34.73 & 1.66 \\
            C2RF  & 0.30  & 0.73  & 1.75  & 0.59  & 32.67 & 2.06 \\
            UniVF  & \bgsecond{0.35} & \bgsecond{0.76} & \bgsecond{2.00} & \bgsecond{0.68} & \bgsecond{29.61} & \bgsecond{1.30} \\
            MambaVF & \bgfirst{0.37} & \bgfirst{0.78} & \bgfirst{2.00} & \bgfirst{0.69} & \bgfirst{29.51} & \bgfirst{1.26} \\
            \bottomrule
            \end{tabular}%
        }
    \end{minipage}
    \vspace{-1em}
\end{table*}

\begin{table*}[t]
    \centering
    \caption{Results of the ablation experiments, with the best-performing values highlighted in \colorbox{firstcolor}{red}.}
    \label{tab:ablation}%
    \resizebox{\linewidth}{!}{
        \begin{tabular}{lcccccccccc}
            \toprule
            \multicolumn{1}{c}{\multirow{2}{*}{Descriptions}} & \multicolumn{3}{c}{Configurations}   & \multicolumn{6}{c}{Metrics}     \\
            \cmidrule(lr){2-4}\cmidrule(lr){5-10}
            & {Scan type} & {3D Decoder} & {Multi-frame} &  {VIF $\uparrow$} & {SSIM $\uparrow$} & {MI $\uparrow$} & {Qabf $\uparrow$} & {BiSWE $\downarrow$} & {MS2R $\downarrow$} \\
            \midrule
            Exp.~\uppercase\expandafter{\romannumeral1}: 2D spatial scan &   2D    &    \XSolidBrush   &   \Checkmark     &  0.74 & 0.98 & 3.83 & 0.70 & 8.02  & 0.20  \\
            Exp.~\uppercase\expandafter{\romannumeral2}: 1D temporal scan &   1D    &    \XSolidBrush   &    \Checkmark    &  0.62 & 0.95 & 2.97 & 0.58 & 8.35  & 0.17  \\
            Exp.~\uppercase\expandafter{\romannumeral3}: w/o state interaction & 3D & \XSolidBrush & \Checkmark & 0.79 & 0.99 & 4.16 & 0.73 & 7.87  & 0.16\\
            Exp.~\uppercase\expandafter{\romannumeral4}: Single-frame input &   2D    &  \XSolidBrush     &  \XSolidBrush     &  0.68 & 0.98 & 3.49 & 0.70 & 7.94  & 0.18  \\
            Exp.~\uppercase\expandafter{\romannumeral5}: 3D Decoder &  3D    &   \Checkmark      &  \Checkmark     &  0.79 & 0.99 & 3.86 & 0.73 & 7.68  & 0.16 \\
            \midrule
            MambaVF (Ours)  &  3D    &   \XSolidBrush     &  \Checkmark    & \bgfirst{0.82} & \bgfirst{0.99} & \bgfirst{4.45} & \bgfirst{0.75} & \bgfirst{7.47} & \bgfirst{0.15} \\
            \bottomrule
    \end{tabular}}
\end{table*}%
\begin{table*}[!t]
    \centering
    \begin{minipage}[t]{0.46\textwidth}
        \vspace{0pt}
        \centering
        \captionof{table}{Efficiency comparison among video fusion methods\protect\footnotemark. Red and blue backgrounds mark the lowest and second-lowest values.}
        \label{tab:params_results}
        \resizebox{\linewidth}{!}{%
        \begin{tabular}{ccccc}
            \toprule
            Methods & Task & Params (M) & FLOPs (G) & Latency (ms)\\
            \midrule
            TemCoCo & IVF & 19.18  & 158.54   & \bgfirst{{12.70}}\\
            VideoFusion & IVF & \bgsecond{{6.74}} & 85.69 & 18.93\\
            SDMFusion & IVF & 835.86 & 244.99   & 94.35\\
            DRFusion & IVF & 86.62  & 35299.93 & 1236.15\\\cdashline{1-5}[1pt/1pt]
            UniVF & Unified & 9.15 & \bgsecond{{81.48}} & 26.72\\
            MambaVF & Unified & \bgfirst{{0.52}} & \bgfirst{{10.92}} & \bgsecond{{12.91}}\\
            \bottomrule
        \end{tabular}}
    \end{minipage}
    \hfill
    \begin{minipage}[t]{0.53\textwidth}
        \vspace{0pt}
        \centering
        \includegraphics[width=\linewidth]{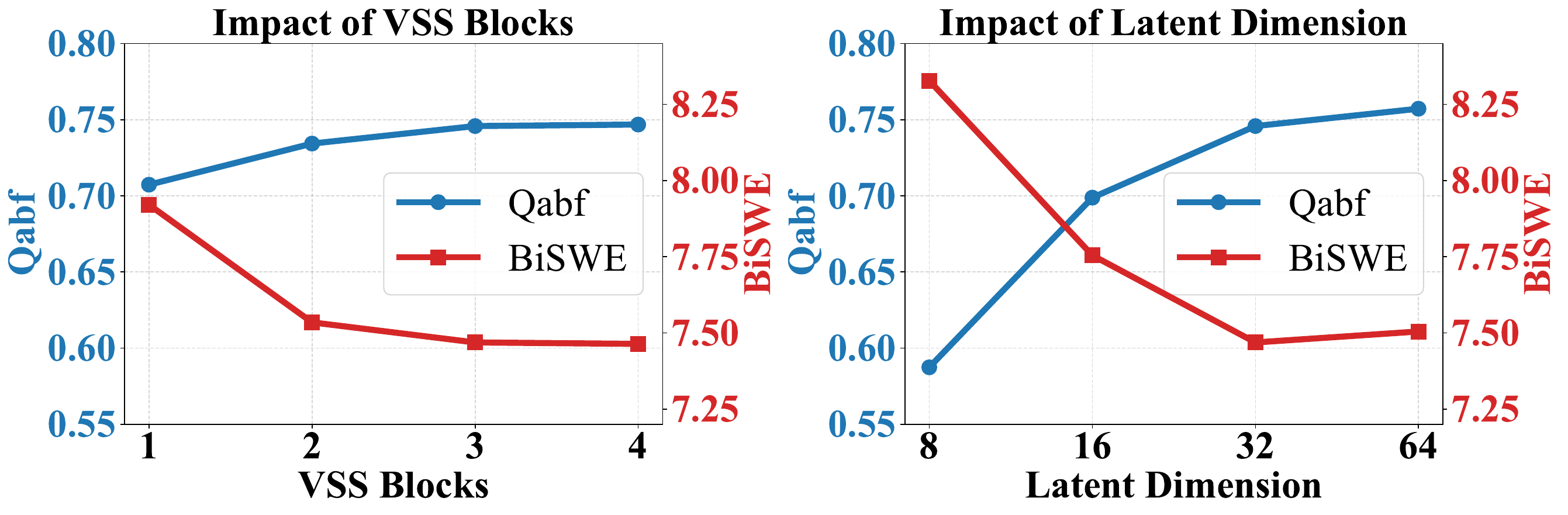}
        \captionof{figure}{Hyperparameter analysis of VSS blocks and latent dimension.}
        \label{fig:Hyper}
    \end{minipage}
\end{table*}

\subsection{Video Fusion Experiments}
We compare MambaVF with SOTA methods on four video fusion tasks to demonstrate its superior performance. Qualitative video comparison results are provided in the supplementary material.

\bfsection{Multi-Exposure Video Fusion}
The MEF task requires the model to synthesize high-dynamic-range content from sequences with varying exposure levels. We compare MambaVF against SOTA methods including 
CUNet~\citep{deng2020CUNET}, HoLoCo~\citep{DBLP:HoLoCo}, CRMEF~\citep{liu2024searching}, TC-MoA~\citep{DBLP:TCMOA}, FILM~\citep{DBLP:FILM}, ReFusion~\citep{DBLP:ReFusion}, and UniVF~\citep{zhao2025unified}. As illustrated in \cref{fig:MEF}, MambaVF demonstrates a superior ability to recover details in both extremely dark and saturated regions, modeling the temporal evolution of luminance implicitly. This leads to more natural transitions of light and shadow, achieving higher VIF scores and significantly lower MS2R values in \cref{tab:MFFMEF}, proving that MambaVF can maintain high-fidelity textures while ensuring temporal alignment in dynamic lighting conditions.

\bfsection{Multi-Focus Video Fusion}
In MFF, the objective is to integrate the focused regions from different focal planes into an all-in-focus video. We evaluate MambaVF against CUNet~\citep{deng2020CUNET}, RFL~\citep{DBLP:RFL}, EPT~\citep{DBLP:EPT}, TC-MoA~\citep{DBLP:TCMOA}, FILM~\citep{DBLP:FILM}, ReFusion~\citep{DBLP:ReFusion}, and UniVF~\citep{zhao2025unified}. The VF-Bench results in \cref{fig:MFF,tab:MFFMEF} show that MambaVF reduces halo artifacts near focus boundaries by capturing spatial and temporal sharpness cues. Our method accurately identifies the transition zones between focused and defocused areas, preserving sharp edges across the entire sequence. MambaVF outperforms baseline models, particularly in scenes with complex depth-of-field changes where traditional image-based methods struggle to maintain boundary consistency.

\bfsection{Infrared-Visible Video Fusion}
We compare our model with 
EMMA~\citep{DBLP:EMMA}, 
FILM~\citep{DBLP:FILM}, 
Text-DiFuse~\citep{DBLP:Text-Diffuse}, 
DCEvo~\citep{Liu_2025_CVPR_DcEVO}, 
TemCoCo~\citep{gong2025temcoco}, 
VideoFusion~\citep{tang2026videofusion}, 
SDMFusion~\citep{Liu_2026_CVPR}, 
DRFusion~\citep{li2026drfusion}, 
and UniVF~\citep{zhao2025unified}. MambaVF excels at preserving the high-contrast thermal signatures of infrared sources while retaining the rich environmental details of visible light. Qualitatively, our fused videos exhibit more prominent targets and clearer background structures in \cref{fig:IVF}. Quantitatively, in \cref{tab:IVF}, MambaVF achieves leading performance in VIF, SSIM, MI and MS2R, indicating that it successfully integrates the maximum amount of complementary information from both modalities while preserving video fluidity.

\bfsection{Medical Video Fusion}
For the MVF task, we fuse MRI images with CT, PET, or SPECT images. We compare MambaVF with EMMA~\citep{DBLP:EMMA}, FILM~\citep{DBLP:FILM}, ReFusion~\citep{DBLP:ReFusion}, Text-DiFuse~\citep{DBLP:Text-Diffuse}, BSAFusion~\citep{DBLP:BSAFusion},  C2RF~\citep{DBLP:C2RF}, and UniVF~\citep{zhao2025unified}. MambaVF demonstrates exceptional performance in maintaining modality information. It preserves fine-grained tissue textures of MRI images while successfully superimposing the functional intensity distributions from PET or SPECT. The results in \cref{tab:MVF,fig:MVF} show that our model achieves the highest structural similarity scores and interframe similarity, effectively capturing complex spatial relationships within medical volumes.

\subsection{Efficiency Analysis}

A cornerstone of MambaVF is its superior computational efficiency, which is particularly vital for real-time video processing. 
We compare MambaVF with all video fusion competitors, including the \textit{unified} video fusion framework UniVF~\citep{zhao2025unified} and four \textit{IVF-specific} video fusion methods: 
TemCoCo~\citep{gong2025temcoco}, VideoFusion~\citep{tang2026videofusion}, SDMFusion~\citep{Liu_2026_CVPR}, and DRFusion~\citep{li2026drfusion}.

As detailed in \cref{fig:first,tab:params_results}, MambaVF requires only \textbf{5.7\%} of the parameters and \textbf{13.4\%} of the FLOPs compared with the unified fusion baseline UniVF, achieving a \textbf{2.1$\times$} speedup. This drastic reduction in computational footprint stems primarily from our flow-free design, which bypasses the heavy resource requirements of motion estimation and feature warping modules.

Compared with the IVF-specific methods, MambaVF has the lowest reported parameter and FLOP counts while supporting all four fusion tasks. Its inference latency is lower than those of VideoFusion, SDMFusion, and DRFusion, and very close to that of TemCoCo. Thus, MambaVF combines broader task coverage with low reported complexity and competitive latency.

\subsection{Ablation Studies}

To further investigate the impact of our design choices, we conduct several controlled experiments on the MEF task in \cref{tab:ablation}.

\bfsection{Architectural Components}
We evaluate STB Scanning and the MuFuse Module against three restricted baselines: 

(1) Exp. \uppercase\expandafter{\romannumeral1}: 2D spatial scan. Replace STB scanning with a standard 4-way 2D spatial scan;\\
(2) Exp. \uppercase\expandafter{\romannumeral2}: 1D temporal scan. Replace STB scanning with a simplified 1D temporal-only scan;\\
(3) Exp. \uppercase\expandafter{\romannumeral3}: w/o state interaction. {Remove the cross-source terms $C_{2,t}h_t^2$ and $C_{3,t}h_t^1$} in \cref{eq:mufuse_readout}, while retaining each source's independent Mamba scan and the full STB Scanning schedule.

The quantitative results show that 2D spatial scanning cannot model inter-frame dependencies, leading to higher BiSWE and visible flickering, while 1D temporal scanning captures motion but sacrifices fine-grained spatial structures. In contrast, our 8-way STB scanning jointly captures pixel evolution and intra-frame geometry from multiple perspectives, achieving a better spatiotemporal balance. Removing reciprocal state interaction consistently degrades performance, demonstrating that exchanging source-specific state information improves both information preservation and temporal consistency beyond independent source modeling.

\footnotetext{All compared methods are video fusion methods;
image fusion methods are not included.}

\bfsection{Temporal Context and Decoder Refinement}
We further examine temporal context and reconstruction through two variants:

(4) Exp. \uppercase\expandafter{\romannumeral4}: Single-frame input. Use only the target frame from each source, removing neighboring-frame context and degrading MambaVF into a frame-by-frame image fusion model.\\
(5) Exp. \uppercase\expandafter{\romannumeral5}: 3D Decoder ResBlocks. Replace the decoder's 2D residual blocks with their 3D counterparts to assess additional spatio-temporal processing during reconstruction.

The transition from multi-frame to single-frame input (Exp. \uppercase\expandafter{\romannumeral4}) caused a substantial drop in overall fusion quality, underscoring the necessity of temporal context for maintaining coherence in video fusion. Regarding the decoder architecture (Exp. \uppercase\expandafter{\romannumeral5}), while 3D Residual Blocks aim to capture deeper spatio-temporal features, they actually yielded a slight performance decrease and introduced significant computational overhead. Adhering to the principle of parsimony, we retain the 2D Residual Blocks, as they provide superior texture refinement with significantly lower resource consumption.

\bfsection{Impact of VSS Blocks and Hidden State Dimension}
We analyze the sensitivity to the number of VSS blocks and the latent dimension in \cref{fig:Hyper}. Increasing the depth of the Mamba encoder improves the fusion performance but introduces additional latency. We found that a depth of 3 blocks provides an optimal trade-off for most real-time scenarios. 
Furthermore, increasing the latent dimension beyond 32 does not lead to significant performance gains, but it does increase computational cost. Therefore, we choose 3 VSS blocks with an embedding dimension of 32 as our configuration.

\section{Conclusion}
\label{sec:con}

We introduced MambaVF, an efficient and flow-free framework for multi-source video fusion based on state space models. By shifting away from the traditional estimate-and-warp paradigm, we address the long-standing bottlenecks of computational redundancy and flow-dependency in video fusion. 
In MambaVF, the MuFuse Mamba Encoder couples source-specific context extraction with reciprocal state interaction, while eight-direction STB Scanning aggregates complementary spatial and temporal information without explicit motion estimation or feature warping. 
Experimental results on VF-Bench show that MambaVF not only achieves SOTA performance across four fusion tasks but also offers a significant reduction in parameters, FLOPs and runtime. We believe MambaVF provides a new, high-performance baseline for the deployment of fusion algorithms on resource-constrained platforms and offers a promising direction for future research in efficient spatio-temporal modeling and fusion.

\subsection*{AI use statement}

In this work, we used generative AI tools solely to improve the grammar, clarity, and readability of text written by the authors. Generative AI was not used to generate research ideas, formulate hypotheses or claims, design the methodology or experiments, analyze data, or interpret results. The authors reviewed all AI-assisted edits and take full responsibility for the final content of this work.

\bibliography{iclr2027_conference}
\bibliographystyle{iclr2027_conference}

\newpage
\appendix
\section{Additional Qualitative Fusion Comparison Results}\label{sec:sm_fusionimage}
\begin{figure}[h]
    \centering
    \includegraphics[width=\linewidth]{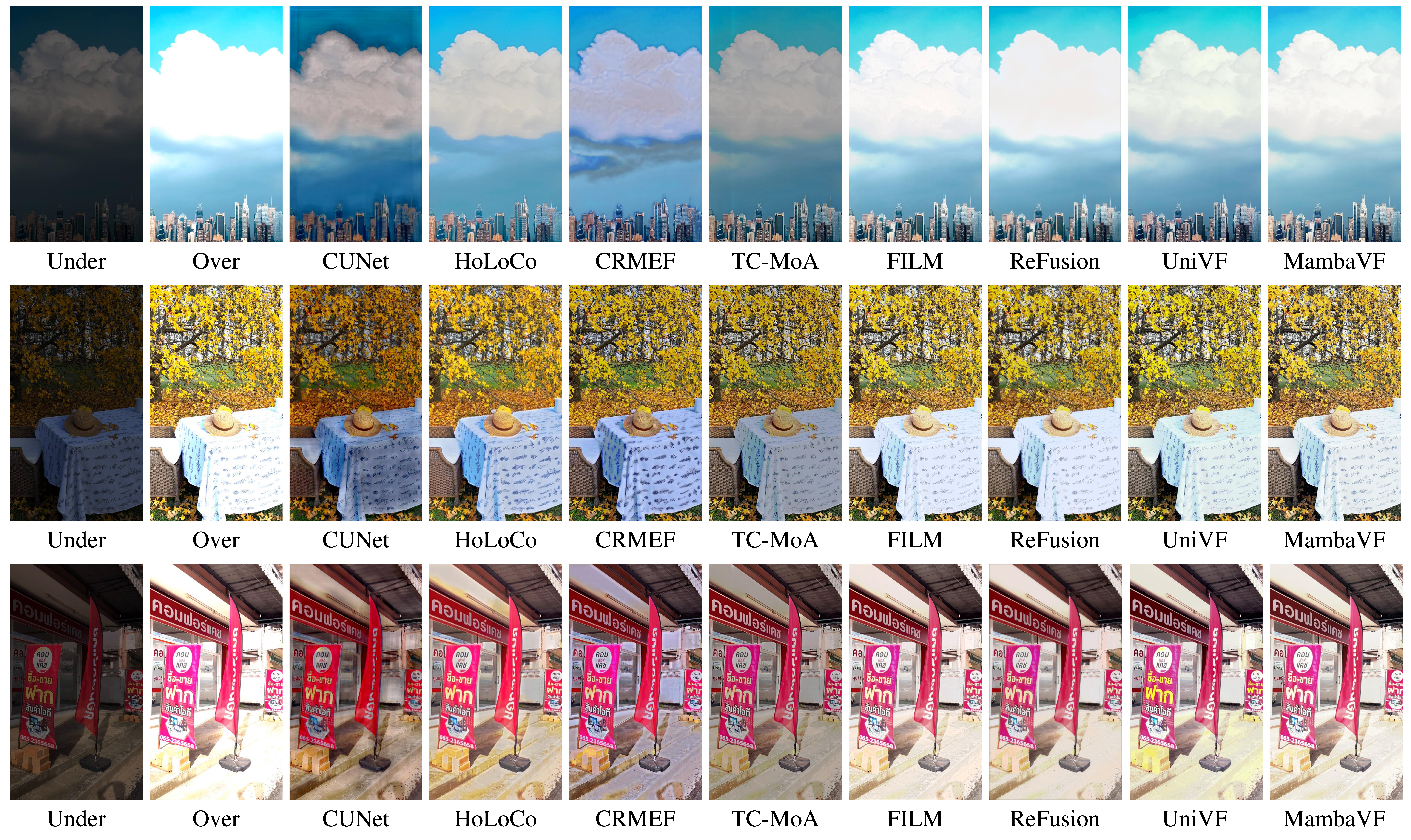}
    \caption{Visual comparison of fused results for multi-exposure video fusion.}
    \label{fig:SM_MEF1}
\end{figure}

\begin{figure}[t]
    \centering
    \includegraphics[width=\linewidth]{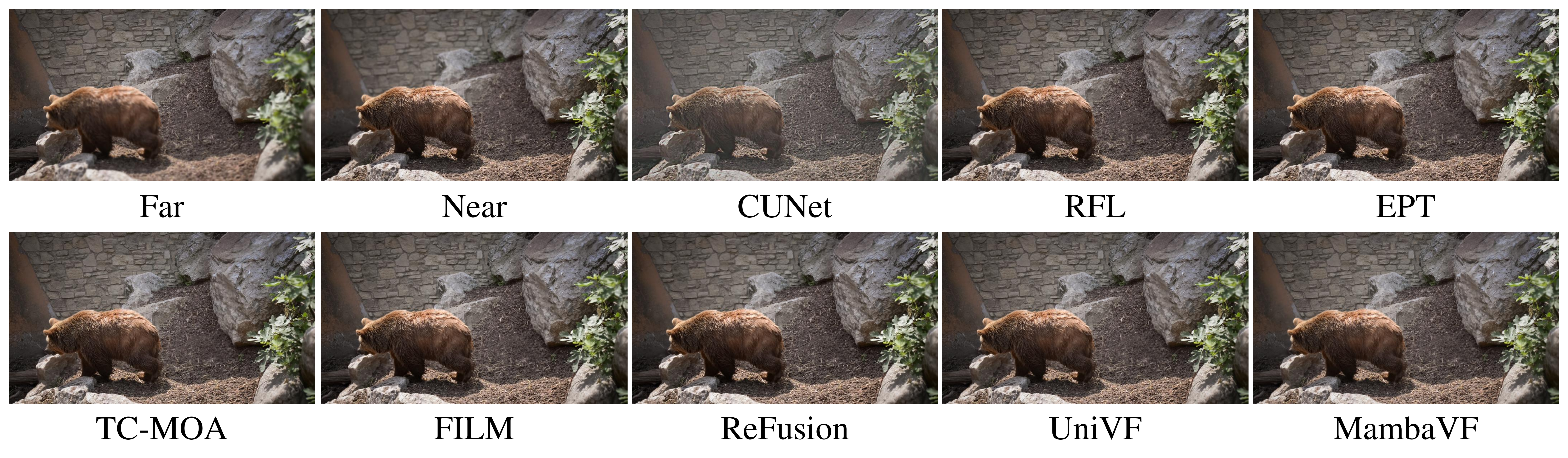}
    \caption{Visual comparison of fused results for multi-focus video fusion.}
    \label{fig:SM_MFF1}
\end{figure}

\begin{figure}[t]
    \centering
    \includegraphics[width=\linewidth]{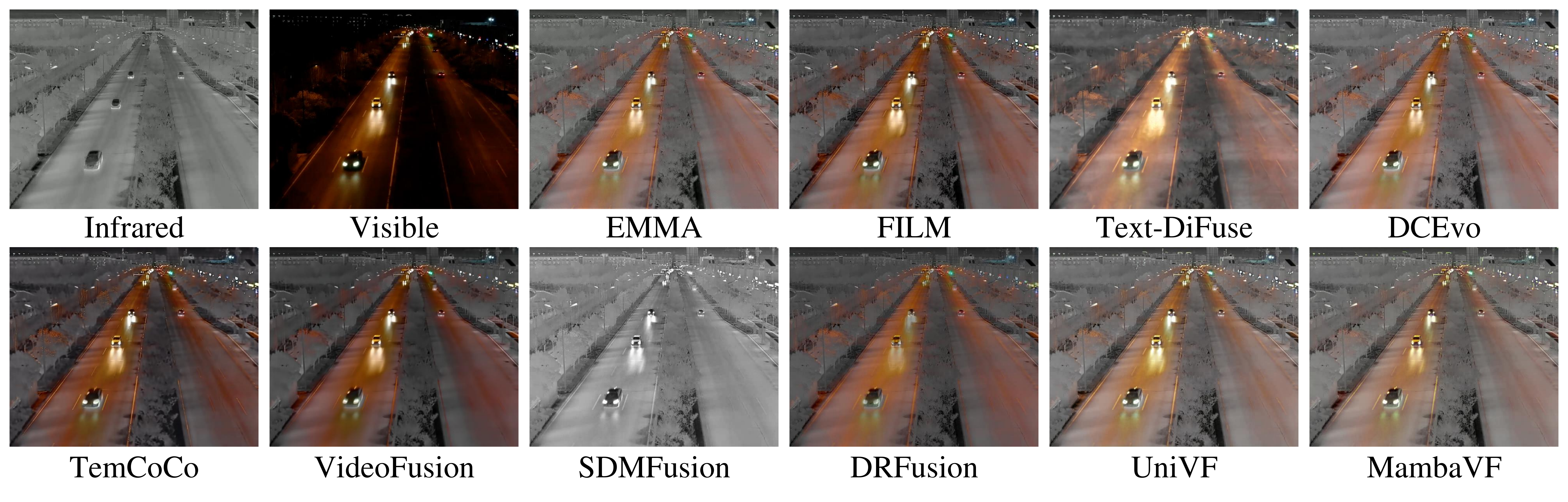}
    \caption{Visual comparison of fused results for infrared-visible video fusion.}
    \label{fig:SM_IVF1}
\end{figure}

\begin{figure}[t]
    \centering
    \includegraphics[width=\linewidth]{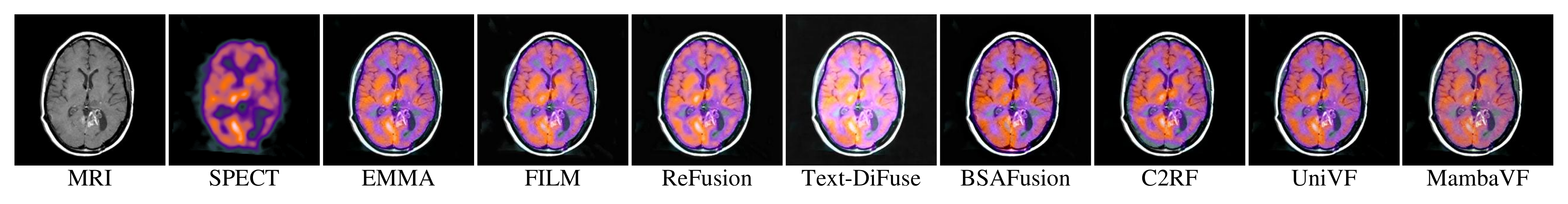}
    \caption{Visual comparison of fused results for medical video fusion.}
    \label{fig:SM_MIF1}
\end{figure}

Additional visual evidence is provided in the figures below to further demonstrate the performance of our framework:
\begin{itemize}[left=1pt,itemsep=0pt, topsep=0pt, parsep=0pt]
    \item Further visual results for the \textit{Multi-exposure Video Fusion} task are presented in \cref{fig:SM_MEF1}.
    \item Further visual results for the \textit{Multi-focus Video Fusion} task are presented in \cref{fig:SM_MFF1}.
    \item Further visual results for the \textit{Infrared-Visible Video Fusion} task are presented in \cref{fig:SM_IVF1}.
    \item Further visual results for the \textit{Medical Video Fusion} task are presented in \cref{fig:SM_MIF1}.
\end{itemize}
These visual evaluations corroborate our primary findings: MambaVF faithfully recovers subtle structures and intricate textures from source inputs while successfully merging complementary features into high-quality, information-rich fused frames. Furthermore, the video sequences included in the supplementary materials highlight the robust temporal stability of our method, characterized by a significant reduction in inter-frame jitter and motion artifacts.

\end{document}